\documentclass[10pt,twocolumn,letterpaper]{article}

\usepackage{wacv}
\usepackage{times}
\usepackage{epsfig}
\usepackage{graphicx}
\usepackage{amsmath}
\usepackage{amssymb}
\usepackage{booktabs}
\usepackage{amsfonts}
\usepackage{amssymb}

\usepackage{mathrsfs}
\usepackage{color}
\usepackage{epstopdf}
\usepackage{algorithm}
\usepackage[noend]{algorithmic}
\usepackage{url}
\usepackage{dsfont}
\usepackage{epstopdf}
\usepackage{subfigure}
\usepackage{mathtools}
\usepackage{multirow}
\usepackage{footnote}
\usepackage{gensymb}
\def\wacvPaperID{165} % Enter the WACV Paper ID here

\wacvalgorithmstrack   % Uncomment this line if you are submitting to the Algorithms Track.
\wacvfinalcopy % *** Uncomment this line for the final submission

\ifwacvfinal
\usepackage[breaklinks=true,bookmarks=false]{hyperref}
\else
\usepackage[pagebackref=true,breaklinks=true,colorlinks,bookmarks=false]{hyperref}
\fi

\begin{document}

%%%%%%%%% TITLE
\title{X-NeRF: Explicit Neural Radiance Field for \\Multi-Scene 360$^{\circ} $ Insufficient RGB-D Views}

\author{Haoyi Zhu, Hao-Shu Fang, Cewu Lu\\
Shanghai Jiao Tong University\\
Shanghai, China\\
{\tt\small \{hyizhu1108, fhaoshu\}@gmail.com, lucewu@sjtu.edu.cn}
}

\maketitle
\thispagestyle{empty}

%%%%%%%%% ABSTRACT
\begin{abstract}
   Neural Radiance Fields (NeRFs), despite their outstanding performance on novel view synthesis, often need dense input views. Many papers train one model for each scene respectively and few of them explore incorporating multi-modal data into this problem. In this paper, we focus on a rarely discussed but important setting: can we train one model that can represent multiple scenes, with 360$^\circ $ insufficient views and RGB-D images? We refer insufficient views to few extremely sparse and almost non-overlapping views. To deal with it, X-NeRF, a fully explicit approach which learns a general scene completion process instead of a coordinate-based mapping, is proposed. Given a few insufficient RGB-D input views, X-NeRF first transforms them to a sparse point cloud tensor and then applies a 3D sparse generative Convolutional Neural Network (CNN) to complete it to an explicit radiance field whose volumetric rendering can be conducted fast without running networks during inference. To avoid overfitting, besides common rendering loss, we apply perceptual loss as well as view augmentation through random rotation on point clouds. The proposed methodology significantly out-performs previous implicit methods in our setting, indicating the great potential of proposed problem and approach. Codes and data are available at \url{https://github.com/HaoyiZhu/XNeRF}.
\end{abstract}

%%%%%%%%% BODY TEXT
\section{Introduction}
\begin{figure}[!tb]
\begin{center}
   \includegraphics[width=\linewidth]{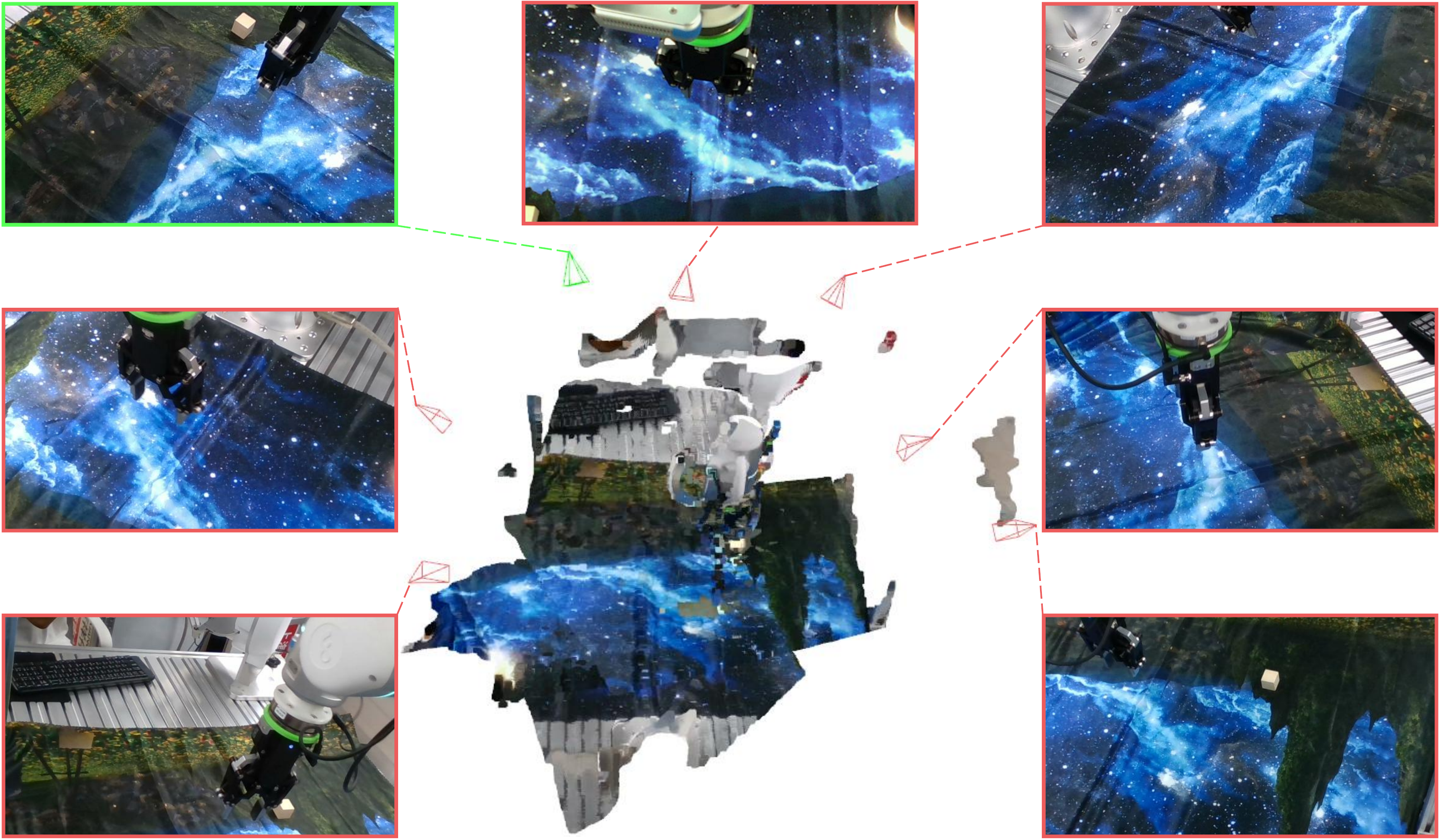}
\end{center}
   \caption{\textbf{An illustration of our problem setting.} The center shows an incomplete scene captured by a few low-cost RGB-D cameras. The small square cones around the scene represent the locations and directions of cameras and their corresponding RGB images are shown in surrounding rectangles. Among them, the red are seen views for training while the green one denotes the unseen view for testing. The insufficient views are very sparse with less than 10\% to 20\% overlapping with each other, making the problem extremely hard.}
\vspace{-0.1in}
\label{fig:insufficient_views}
\end{figure}

Neural Radiance Fields (NeRFs)~\cite{nerf} have aroused significant research interest recently, which usually implicitly encode scenes using coordinate-based multi-layer perceptrons (MLPs) and have a wide range of applications such as novel view synthesis~\cite{mip_nerf,dsnerf,nerf, ibrnet,pixel_nerf,nerf++}.
%, 3D-aware image editing~\cite{editing2, editing1, editing3}, lighting~\cite{lighting1, lighting2, lighting3, lighting4}, pose estimation~\cite{pose4,pose2,pose1,pose3}, etc. 
A lot of follow-up work makes efforts to improve and extend NeRF~\cite{nerf} in various ways from convergence and rendering speed~\cite{dsnerf,fast_nerf,nsvf,dvgo} to dynamic scenes~\cite{dynamic3,dynamic1, dynamic2}, etc. Some methods~\cite{dvgo,plenoxels,plenoctrees} utilize explicit structures to gain huge performance improvement but they still directly encode scenes in learnable network parameters.

Despite the exceptional performance in lots of scenarios, most of NeRF-like methods need a lot of densely captured views when training, making them hard or expensive to apply to practice. Although some work~\cite{dsnerf, pixel_nerf} has studied few-view training, their usual applicable scenarios require views with small perspective changes and large overlapping. What's more, most methods usually train a model for only one scene given the implicit modeling, making it difficult to apply them to massive scenes. Finally, with the rapid development of hardwares, depth data is increasingly available. But most current NeRF-related work only take RGB modality as input. How to utilize the depth information for better rendering deserves more exploration. %Therefore, it is necessary to expand novel view synthesis task to RGB-D multi-modal scenario. 

To this end, in this paper we aim to propose a methodology which allows a single model to (i) deal with multiple scenes, (ii) with insufficient views that are 360$^{\circ} $ around the scenes and (iii) incorporate the depth data for better rendering. Fig.~\ref{fig:insufficient_views} displays an example of our setting. % Generally speaking, it might be an ill-posed problem given only a few insufficient views 360$^{\circ} $ around, but with the depth information, the problem makes sense and is totally solvable.
% Based on the above, in this paper we discuss an extreme and challenging setting where the seen views are from multi-scene, multi-modal, 360$^{\circ} $, and insufficient. Specifically, we say a set of views is `insufficient' when they are extremely sparse and have almost no overlapping. This is a meaningful problem because in many low-cost situations we want to train one model for multiple scenes while we can only be access to or be able to afford insufficient views. And if without depth information, the problem is likely to be ill-posed.
To tackle this hard setting, we propose an explicit neural radiance field (X-NeRF), which can take RGB-D images as inputs. Different from other NeRF-like approaches implicitly mapping coordinates to colors and densities, we explicitly modeling this problem as a completion task. The intuition behind comes from the observation that given a few RGB-D input images, a large part of the scene is actually known. In other words, plenty of information is already available initially, so that we only have to learn a general scene-irrelevant completion relation. Since the network is designed to encode a general completion mapping rather than a specific scene, we can naturally deal with the multi-scene problem. 

Specifically, the input RGB-D images are converted to sparse colorful point clouds and quantized to sparse tensors on which we can directly apply Minkowski Engine~\cite{me} to operate. We adopt a 3D sparse generative CNN to construct and complete the explicit neural radiance fields. Our backbone applies a UNet-like~\cite{unet} encoder-decoder structure with multi-stage generative transposed convolution and pruning layers in the decoder. To avoid overfitting on seen views, besides common rendering loss, we also apply perceptual loss with patch-wise sampling as well as view augmentation through random rotation on point clouds.
% In the encoder, the convolution operations only act on non-zero coordinates so that no new coordinates are generated, while in the decoder, generative transposed convolutions play a role to up-sample and generate new coordinates. To stabilize and accelerate the training process, the decoder has multiple output stages corresponding to different resolutions and each stage's output radiance field participates in loss calculation and back propagation. Synchronously, pruning operation is applied in each stage to removing those coordinates whose corresponding density is too small and location is too far away from input point cloud to reduce memory cost. 
Volumetric rendering with post-activation is used. By shooting and querying a ray from a pixel 
% the colors and densities along the ray
, the accumulated color and depth of it can be rendered.

Extensive experiments demonstrate that the proposed task is extremely challenging for existing methods while our approach can handle it well. % Since the problem of multi-scene 360$^\circ $ insufficient RGB-D views has been discussed rarely by previous methods, 
We first compare our approach with DS-NeRF~\cite{dsnerf}, an advanced NeRF-based work that also supports depth supervision, and DVGO~\cite{dvgo} which is a state-of-the-art NeRF-like method utilizing explicit structures, on single scene experiments. To be fair, we add depth supervision to DVGO~\cite{dvgo}. Then we compare X-NeRF with some recent NeRF-related work that supports multi-scene training such as pixelNeRF~\cite{pixel_nerf} and IBRNet~\cite{ibrnet} (depth supervision is also added). The results state clearly that X-NeRF is robust with multi-scene 360$^\circ $ insufficient views and can produce reliable novel view predictions. Our work outperforms previous methods on the extreme setting, indicating that X-NeRF can be applied to practice in a low-cost manner as we can train one model for many scenes while the inference process is quite lightweight. % The multi-scene success also indicates that X-NeRF can be applied to practice in a low-cost manner as we can train one model for many scenes while the inference process is quite lightweight, showing that our methodology is promising.

\section{Related Work}
{\bf Novel view synthesis.}
To synthesize a novel view image given a set of images is a classic and long-standing task. Rendering methods can be mainly divided into image-based or model-based. Image-based methods~\cite{image_based1,image_based2,ibrnet} directly learn the transformation on image level such as warping or interpolation, which are typically more computational efficient. However, they need reference views during inference and the number and density of reference images may influence the rendering quality greatly. Model-based methods~\cite{model_based2,model_based3,model_based4,model_based1,model_based5} express scenes as high dimensional representations and apply physically meaningful models such as optical model~\cite{optical_model} to render the novel view images. There are various forms to represent scenes. Earlier works apply lumigraph~\cite{lumigraph1,lumigraph2} and light fields~\cite{lf1,lf2,lf3,lf4} to directly interpolate on input images. Nevertheless, they need exceedingly dense inputs which is totally unaffordable in many applications. Other methods utilize explicit representations such as mesh~\cite{mesh1,mesh2,mesh3,mesh4} to deal with sparse inputs. However, mesh-based approaches cannot work well with gradient-based optimization due to discontinuities and local minima. Recently, many deep learning based methods employ CNNs to construct multi-plane images (MPIs)~\cite{MPI1,MPI2,MPI3,MPI4,MPI5,MPI6} for forward-facing captures. There are also approaches that encode scenes as volumetric representations~\cite{model_based2,model_based3,MPI3,MPI5,model_based5,MPI6}, but they often struggles with complex and large-scale scenes.

{\bf Neural Radiance Fields.}
NeRFs have aroused great interest and achieved huge success in novel view synthesis task in recent years. A classic NeRF~\cite{nerf} learns a direct mapping from coordinates to corresponding textures such as color and density, implicitly encoding a scene in MLPs. Since proposed, people have extended NeRF~\cite{nerf} to a lot of variants with different characteristics including editable~\cite{editing2,editing1,editing3}, fast inference and/or training~\cite{dsnerf,fast_nerf,nsvf,dvgo}, deformable~\cite{deform2,deform1}, unconstrained images~\cite{unconst2,unconst1}, etc. 
Some recent work~\cite{dvgo,plenoxels,plenoctrees} introduces explicit structures to gain great performance enhancement, which indicates that the implicit MLP architecture is not necessarily the key to success. Nevertheless, despite the explicit voxel grid structures, they are actually still essentially an implicit modeling, as they still encode the scene information in learnable parameters. The implicit modeling makes NeRF-based methods hard to freely generalize on multi-scene cases. Though some work such as~\cite{pixel_nerf} claims that they have the ability to deal with multi-scene task, they in practice can only process multiple small objects or multiple similar simulated scenes. Moreover, when it comes to the extreme situation proposed in this paper that the input views are insufficient, which means the input views are extremely sparse but 360$^{\circ} $ around the real scenes and have almost no overlapping (often less than 10\% to 20\%), the implicit modeling easily overfits to a trivial solution due to its little constraints on the scene structure. Some approaches can deal with few inputs such as \cite{dsnerf, pixel_nerf}, but their applicable scenario is mostly forward-facing captures which is actually still not sparse enough. % We propose a novel fully explicit modeling in this paper which can deal with the challenging insufficient-view task excellently.

{\bf Multi-modal RGB-D data.} Nowadays, with the rapid development of hardware devices, depth modal is becoming increasingly common and available. It is often much more affordable and cheaper to capture few insufficient RGB-D views than to capture tens or hundreds times more RGB views as long as we tolerate some reasonable depth errors.\footnote{\href{https://www.jobted.com/salary}{The average salary in U.S.} reaches \$53,490 per year and \href{https://www.amazon.com/Intel-Realsense-D435-Webcam-FPS/dp/B07BLS5477}{the price of an Intel RealSense RGB-D camera} is only \$297.17.} Therefore, it is significantly important to process RGB-D inputs and accurately render novel view depth images. Synthesizing perfect novel views from only insufficient 360$^{\circ} $ RGB images is probably ill-posed, but with the additional depth modal, the problem is wholly solvable. Deng et al.~\cite{dsnerf} already make use of depth information, and shows depth knowledge can greatly avoid overfitting on seen views as well as benefit the convergence and performance of NeRFs.

\begin{figure*}[ht]
  \centering
  \includegraphics[width=0.95\linewidth]{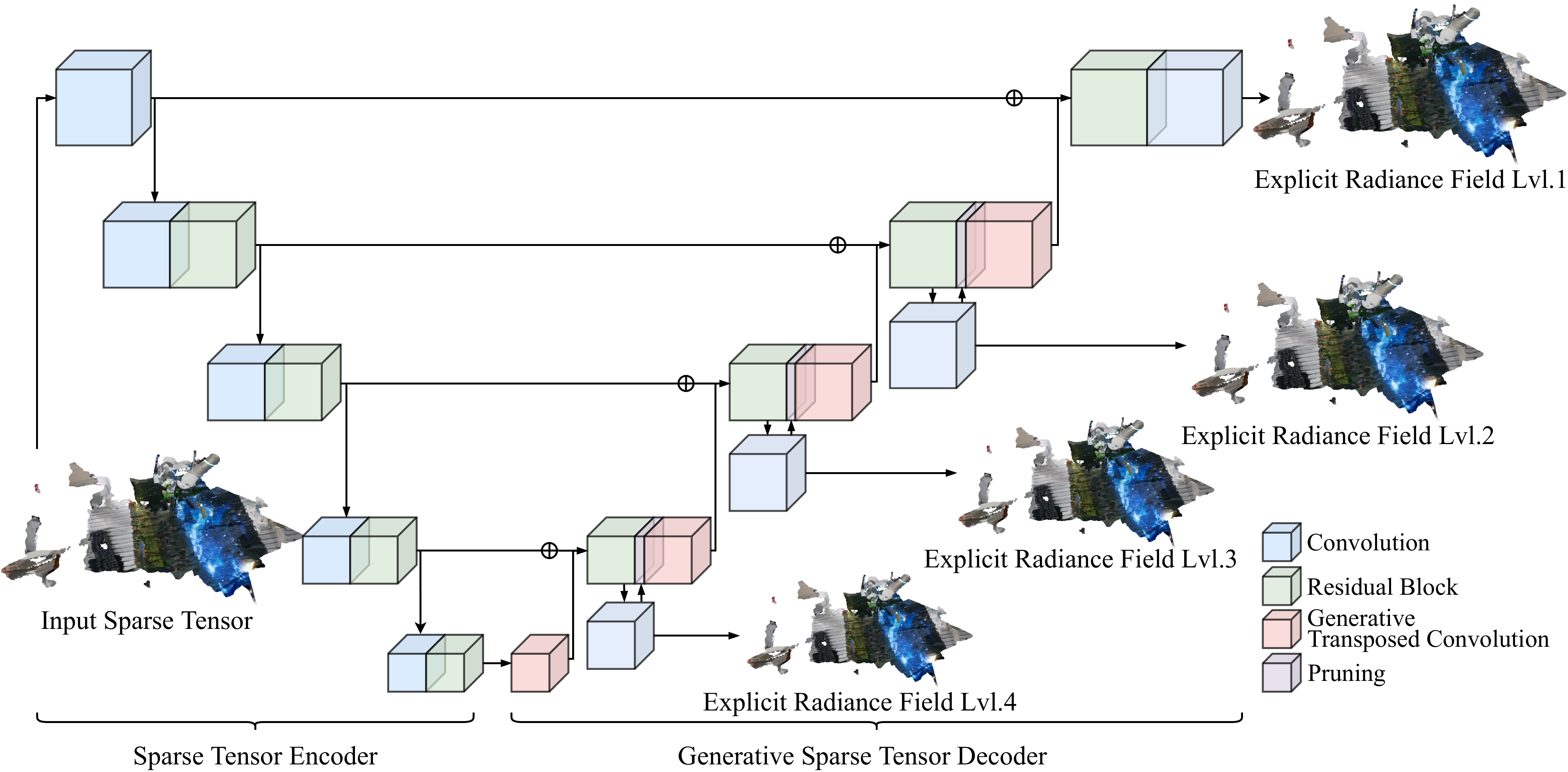}
  \caption{\textbf{Overview model architecture of X-NeRF.} We use a 3D sparse generative convolutional neural network to accomplish our fully explicit completion modeling. Given a sparse tensor representing a specific incomplete scene, we use a encoder-decoder structure similar to UNet to complete and map it to an explicit neural radiance field. The encoder only operates on existing coordinates to save cost when extracting features. Then in the multi-stage decoder, generative transposed convolutional layers with pruning layers are applied to produce novel points. Each stage gives an output with different resolution, helping to stabilize the training process.}
  \label{fig:model_architecture}
\end{figure*}

\section{Preliminaries}
\label{sec:preliminaries}
%\subsection{Volumetric Rendering}
%\label{sec:pre_vol_render}
NeRF-based methods take as inputs a set of images of different views and implicitly map 3D coordinates to densities $\mathbf{\sigma}$ and colors $\mathbf{c}$, encoding a particular scene into network parameters: $f(\mathbf{x},\mathbf{d})=(\mathbf{\sigma}, \mathbf{c})$. Usually $\operatorname{Sigmoid}$ is acted on $\mathbf{c}$ and $\operatorname{ReLU}$ or $\operatorname{Softplus}$ is acted on $\mathbf{\sigma}$.

Given a particular camera pose $P$, to render the corresponding 2D image pixels, we first emit rays $r$ in the direction $d$ from the projection center $o$ of the camera to the pixels. Then $N$ ordered query points on $r$ between the pre-defined near and far planes are sampled and fed into model to obtain their densities and colors $\{(\sigma_i, \mathbf{c}_i)_{i=1}^{N}\}$, so that we can integrate them using the optical model proposed by \cite{optical_model} to derive the rendered pixel color $\hat{C}(r)$:
\begin{subequations}
\begin{align}
\alpha_{i} &=1-\exp \left(-\sigma_{i} \delta_{i}\right)\; \; 1\leq i \leq N\text{ ,} \label{eq:raw2alpha}\\
T_{i} &=\prod_{j=1}^{i-1}\left(1-\alpha_{j}\right)\; \; 1\leq i \leq N\text{ ,} \label{eq:alpha2weight}\\
\hat{C}(r) &=\left(\sum_{i=1}^{N} T_{i} \alpha_{i} c_{i}\right)+T_{\mathrm{K+1}} c_{\mathrm{bg}}\text{ ,} \label{eq:render_color}
\end{align}
\label{eq:vol_render}
\end{subequations}
where $\alpha_{i}$ represents the termination probability at point $i$ and the accumulated transmittance to point $i$ is denoted by $T_i$. $\delta_{i}$ is the sampling step size, i.e., the distance to the adjacent sampled point on a ray. $c_{\mathrm{bg}}$ is a pre-defined background color, usually either 0 or 1. Depth rendering is similar to color rendering, which can be given by:
\begin{equation}
    \hat{D}(r) =\sum_{i=1}^{N} T_{i} \alpha_{i} d_i\text{ ,} \label{eq:render_depth}
\end{equation}
where $d_i$ is the distance from the ray's origin to point $i$.

\section{Method}
\label{sec:method}

In this section, we introduce Explicit Neural Radiance Field (X-NeRF), an explicit representation for novel view synthesis from multi-scene 360$^{\circ} $ insufficient-view RGB-D images. Instead of implicitly constructing scenes in neural network parameters, we consider a fully explicit methodology using a 3D sparse generative CNN to learn a general scene-irrelevant completion relationship. In the following, we first describe our explicit modeling methodology (Sec.~\ref{sec:explicit_modeling}), then the detailed model architecture of X-NeRF (Sec.~\ref{sec:model_architecture}) and finally our volumetric rendering process (Sec.~\ref{sec:volumetric_rendering}) as well as optimization functions (Sec.~\ref{sec:optimization}).

\subsection{Explicit Modeling}
\label{sec:explicit_modeling}
Existing NeRF-like methods are all essentially implicit, though some such as~\cite{dvgo, plenoxels,nsvf} take use of explicit structures. In spite of their promising performance in many situations, implicit models struggle with three challenges. The first is that when there are insufficient, i.e. extremely sparse and almost non-overlapping, seen views, they tend to overfit to a trivial solution since they have no constraints nor priors on the scene structures. The second is that it is hard for them to naturally process multiple scenes using one model as they directly encode the scene information in model parameters. A few existing NeRF-based methods that support multi-scene learning either need reference views~\cite{ibrnet,pixel_nerf} whose number and density have impact on the rendering effect, or employ independent explicit structures before a shared MLP~\cite{nsvf} whose space and time costs linearly increase with the rise of scene number. The third is that they usually lack the ability to fuse RGB with depth images which are increasingly popular and available at present.

To this end, we propose a fully explicit approach that can tackle the challenging problem of novel view synthesis from multi-scene 360$^{\circ} $ insufficient seen views, and can naturally fuse RGB and depth modals. We consider the problem as an explicit completion task motivated by the fact that given a few RGB-D views around, we can easily get a large part of the location and color information of points in a specific space, and what we need to do is to complete the whole space, i.e. the explicit neural radiance field. Therefore, our network can be modeled as a completion mapping function:
\begin{equation}
\begin{aligned}
\label{eq:modeling}
&f_{\theta}:\left\{{\mathbf{x}_\mathrm{in}}_{i}, {\mathbf{c}_\mathrm{in}}_{i}\right\}_{i=1}^{i=N} \rightarrow\left\{{\mathbf{x}_\mathrm{out}}_{j}, \mathbf{k}_{j}, \sigma_{j}\right\}_{j=1}^{j=M}\text{ ,}\\
&\text {where } \mathbf{k}=\left(k_{\ell}^{m}\right)_{\ell: 0 \leq \ell \leq \ell_{\operatorname{max}}}^{m:-\ell \leq m}\text{ .}
\end{aligned}
\end{equation}
Here $f_\theta$ represents our neural network with learnable parameters $\theta$, which is a completion mapping from $N$ input RGB-D points consisting of input point cloud coordinates $\{{\mathbf{x}_\mathrm{in}}_i \in \mathbb{R}^{3}\}_{i=1}^{i=N}$ and colors $\{{\mathbf{c}_\mathrm{in}}_i \in \mathbb{R}^{3}\}_{i=1}^{i=N}$, to $M$ explicit neural radiance field points where each $\sigma \in \mathbb{R}$ denotes a scalar opacity while we apply $\mathbf{k}$, a vector of spherical harmonic (SH) coefficients, to express output color information similar to~\cite{plenoxels,plenoctrees}. Each $k_{\ell}^{m} \in \mathbb{R}^{3}$ is a set of 3 coefficients for RGB channels. It has been discussed in~\cite{plenoxels} that spherical harmonics of degree 2 is enough, which requires 9 coefficients per color channel for a total of 27 harmonic coefficients per voxel, and we follow their setting. The SHs enable the output colors ${\mathbf{c}_\mathrm{out}}$ to be view-dependent by querying the SH functions $Y_{\ell}^{m}: \mathbb{S}^{2} \mapsto \mathbb{R}$ given its corresponding view direction $\mathbf{d}$:
% where $f_\theta$ represents our neural network with learnable parameters $\theta$ who is a completion mapping from $N$ input RGB-D points to $M$ 5-$D$ points consisting of 3-$D$ coordinates, corresponding color and density. Typically, $M \textgreater N$.
\begin{equation}
\label{eq:sh2color}
{\mathbf{c}_\mathrm{out}}(\mathbf{d} ; \mathbf{k})={\operatorname{Sigmoid}}\left(\sum_{\ell=0}^{\ell_{\max }} \sum_{m=-\ell}^{\ell} k_{\ell}^{m} Y_{\ell}^{m}(\mathbf{d})\right)\text{ .}
\end{equation}
We note here that SHs are vital to novel view synthesis under insufficient-view condition. More detailed discussions are illustrated in Sec.~\ref{sec:view_aug} and Fig.~\ref{fig:percep_rot}.

% Unlike those implicit methods who have to learn the whole scene from scratch, our modeling the problem as a completion task makes full use of the observation that we already know much scene information before training, making our method easier to learn and has a better generalization ability.

\subsection{Model Architecture}
\label{sec:model_architecture}

\begin{figure*}[tb]
  \centering
  \includegraphics[width=0.85\linewidth]{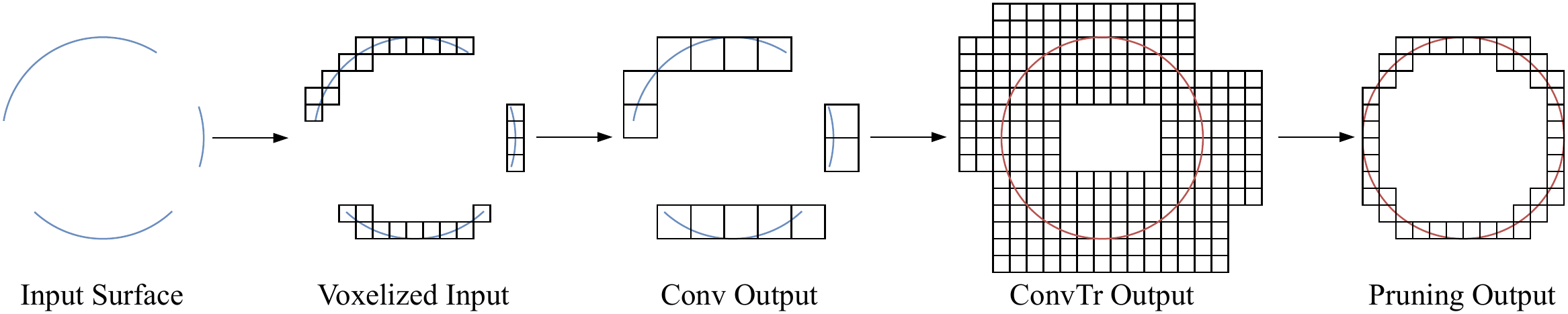}
  \caption{\textbf{A 2-D sketch about the encoding-decoding process.} Given an input surface, we first voxelize it into a sparse tensor which is fed into convolutional encoders. The encoders only operate on existing locations and generate no new coordinates. After that, generative transposed convolutional decoders are applied to up-sample and generate new coordinates so as to complete the scene structure. Finally, optional pruning layers are utilized to remove unnecessary points.}
  \vspace{-0.1in}
  \label{fig:pipeline}
\end{figure*}

Processing RGB-D data is a classic multi-modal problem. In this paper, we convert multi-view RGB-D inputs to colorful point clouds, and voxelize them into Minkowski sparse tensors~\cite{me} so that we can directly apply operations like convolution and transposed convolution on them.

As shown in Fig.~\ref{fig:model_architecture}, we apply an encoder-decoder architecture with skip connections, similar to the structure of UNet~\cite{unet}. The sparse tensor encoder is mainly used for extracting spatial and local features, which is composed of some convolution and residual blocks, while the generative sparse tensor decoder consisting of generative transposed convolution, pruning and residual blocks mainly plays a role of up-sampling and generating new points. The decoder is designed to have multi-stage outputs with increasing resolutions. The outputs of every stage participate in loss computation, which is inspired by the coarse-to-fine design in most of NeRF-related methods.

The encoder only operates on known coordinates. In other words, no new coordinates are generated during encoding. When decoding, we apply generative transposed convolutional layers followed by pruning layers to up-sample and complete the whole radiance field. 
A simple low-dimension schematic diagram of the encoding-decoding pipeline is illustrated in Fig~\ref{fig:pipeline}.
%The overview of the pipeline is shown in Fig~\ref{fig:pipeline}. 
As we can see, the function of pruning is to remove redundant points % generated by generative transposed convolutions 
to save computational resources as well as make the output more accurate. Details for generative transposed convolution layer can be found in~\cite{me,sparse_detection}.

The multi-stage design of decoder is also beneficial to decide which points to keep or to prune. Specifically, in each stage, we prune a point $P_i$ if its corresponding output termination probability $\alpha_i$ is too small and it is too far away from the input point set:
\begin{equation}
    P_i \text{ is pruned} \operatorname{if} \alpha_i \le \tau_\alpha \operatorname{and} \min_{P_j \in \mathcal{C}_{\mathrm{in}}} \operatorname{dist}(P_i, P_j) \le \tau_{\mathrm{dist}}\text{ ,}
    \label{eq:prune}
\end{equation}
where $\tau_\alpha$ and $\tau_{\mathrm{dist}}$ are two hyper-parameters; $\mathcal{C}_{\mathrm{in}}$ denotes the coordinate set of input point cloud; $\operatorname{dist}(\cdot,\cdot)$ represents the operation that computes the Euclidean distance between two points. The intention of the second distance term is that points too far away are not likely to belong to the complete scene. In the practice, we discover that the pruning operation can reduce memory consumption and has little impact on performance, but it may slightly affect the execution speed, so we leave it as an optional choice.

\subsection{Volumetric Rendering}
\label{sec:volumetric_rendering}
Given a few RGB-D images, we can project and voxelize them into a sparse tensor $\mathscr{T}_{\mathrm{in}}$ consisting of a set of coordinates $\mathcal{C}_{\mathrm{in}}$ and the corresponding features $\mathcal{F}_{\mathrm{in}}$:
\begin{subequations}
\begin{align}
\mathscr{T}_{\mathrm{in}}&=(\mathcal{C}_{\mathrm{in}}, \mathcal{F}_{\mathrm{in}})\text{ ,}\\
\mathcal{C}_{\mathrm{in}}&=\{x_i,y_i,z_i\}_{i=1}^{i=N}\text{ ,}\\
\mathcal{F}_{\mathrm{in}}&=\{{\mathbf{c}_\mathrm{in}}_{i}\}_{i=1}^{i=N}\text{ ,}
\end{align}
\end{subequations}
% \begin{equation}
% % \begin{align}
% \mathscr{T}_{\mathrm{in}}=\left(\mathcal{C}_{\mathrm{in}}=\{x_i,y_i,z_i\}_{i=1}^{i=N},\; \mathcal{F}_{\mathrm{in}}=\{{\mathbf{c}_\mathrm{in}}_{i}\}_{i=1}^{i=N}\right)
% % \end{align}
% \end{equation}
where ${\mathbf{c}_\mathrm{in}}_{i} \in \mathbb{R}^3$ represents the 3-channel RGB colors.

Let's denote our X-NeRF model as $f_\theta$, then we can get an output expanded sparse tensor that contains the information of the completed scene, as discussed in Eq.~\ref{eq:modeling}:
\begin{subequations}
\begin{align}
\mathscr{T}_{\mathrm{out}}&=f_\theta(\mathscr{T}_{\mathrm{in}})=(\mathcal{C}_{\mathrm{out}}, \mathcal{F}_{\mathrm{out}})\text{ ,}\\
\mathcal{C}_{\mathrm{out}}&=\{x_j,y_j,z_j\}_{j=1}^{j=M}\text{ ,}\\
\mathcal{F}_{\mathrm{out}}&=\{{\mathbf{k}}_j, \ddot{\sigma}_j\}_{j=1}^{j=M}\text{ ,}
\end{align}
\end{subequations}
where $\ddot{\sigma}_j \in \mathbb{R}$ denotes the raw density before activation.

Sun et al.~\cite{dvgo} has shown that post-activation, i.e. implementing activation function after interpolation operation, is the best choice for volumetric rendering and we follow this setting.
As discussed in Sec.~\ref{sec:preliminaries}, given view directions $\mathbf{d}$ and shooting rays $\mathbf{r}$ corresponding to specific 2D pixels, we first interpolate $\mathscr{T}_{\mathrm{out}}$ on $\mathbf{r}$ to obtain the SHs $\mathbf{k}$ and raw densities $\ddot{\mathbf{\sigma_r}}$ on them, and then use the shifted 
$\operatorname{Softplus}$ 
mentioned in Mip-NeRF~\cite{mip_nerf} to acquire the corresponding densities:
\begin{equation}
\begin{aligned}
\mathbf{\sigma_r} = \log (1+\exp (\ddot{\mathbf{\sigma_r}}+b))\text{ ,}
\end{aligned}
\end{equation}
where the shift $b$ is a hyper-parameter. The RGB colors on each ray point in radiance field can be gotten through Eq.~\ref{eq:sh2color}. After that, we can apply Eq.~\ref{eq:vol_render} and Eq.~\ref{eq:render_depth} to render the 2D pixel colors $\hat{C}(\mathbf{r})$ and depths $\hat{D}(\mathbf{r})$.

\subsection{Optimization}
\label{sec:optimization}
Since all the operations in our pipeline is differentiable, we can optimize X-NeRF through gradient decent. To combat the overfitting issue arising from insufficient views, loss function composed of the following parts is applied.

{\bf Rendering Loss.}
Given a set of RGB-D inputs and camera poses $\mathbf{P}$, the rendering loss is given by the mean squared error (MSE) between ground-truth and rendered outputs:
\begin{equation}
\begin{aligned}
\mathcal{L}_{\mathrm{render} }&=\mathcal{L}_{\mathrm{color}} + \lambda_D \mathcal{L}_{\mathrm{depth}}\text{ ,}\\
\mathcal{L}_{\mathrm{color}}&=\frac{1}{|\mathcal{R}(\mathbf{P})|}\sum_{r \in \mathcal{R}(\mathbf{P})}\|\hat{C}(r)-C(r)\|_{2}^{2}\text{ ,}\\ 
\mathcal{L}_{\mathrm{depth}}&=\frac{1}{|\mathcal{R}_{\mathrm{VD}}(\mathbf{P})|}\sum_{r \in \mathcal{R}_{\mathrm{VD}}(\mathbf{P})}\|\hat{D}(r)-D(r)\|_{2}^{2}\text{ ,}
\label{eq:render_loss}
\end{aligned}
\end{equation}
where $\mathcal{R}(\mathbf{P})$ is the set of rays of $\mathbf{P}$ while $\mathcal{R}_{\mathrm{VD}}(\mathbf{P}) \subseteq \mathcal{R}(\mathbf{P})$ contains rays with valid depths. $\lambda_D$ is a hyper-parameter.

% {\bf Reconstruction Loss.}
% We also apply a reconstruction loss that directly supervise on the input coordinates:
% \begin{equation}
% \mathcal{L}_{\mathrm{recon}}=\frac{1}{|\mathcal{C}_{\mathrm{in}}|} \left( \sum_{P_i \in \mathcal{C}_{\mathrm{in}}}\|\hat{C}(P_i)-C(P_i)\|_{2}^{2} \right)\text{ ,}
% \end{equation}
% where $\mathcal{C}_{\mathrm{in}}$ is the set of input coordinates, while $\hat{C}(P_i)$ and $C(P_i)$ denote point $P_i$'s ground-truth and rendered color.

\begin{figure}[!tb]
  \centering
  \includegraphics[width=\linewidth]{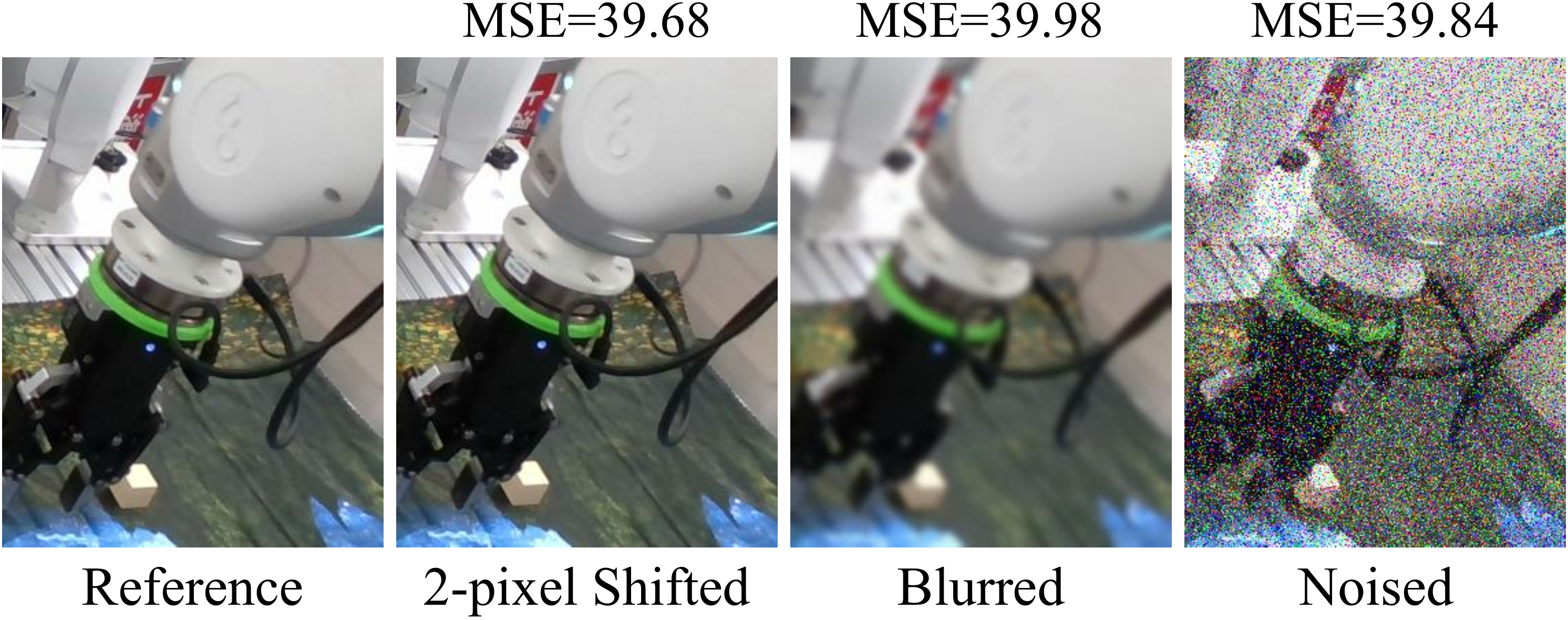}
  \caption{\textbf{Illustrations on the ambiguity of MSE loss.} The three generated examples have similar MSE values, but their perceptual quality varies greatly. From left to right are ground truth, shifted by 2 pixels, Guassian blurred and with random noise.}
%   \vspace{-0.1in}
  \label{fig:mse_examples}
\end{figure}

{\bf Perceptual Loss.}
Besides the vanilla MSE loss, we further add a perceptual loss considering per-pixel MSE error contains no global or high level context information, which may not guide the model to the right road. In practice, lower MSE does not necessarily mean better human perceptual quality. We display intuitive and simple examples in Fig~\ref{fig:mse_examples} for the above two cases. To this end, motivated by many image generation works, we adopt the perceptual loss~\cite{perceptual_loss} to avoid falling into a trivial solution:
\begin{equation}
\mathcal{L}_{\mathrm{percep}} = \sum_{l=1}^{L} \frac{1}{H_{l} W_{l}} \sum_{h, w}\left\|w_{l} \odot\left(\hat{y}_{h w}^{l}-\hat{y}_{0 h w}^{l}\right)\right\|_{2}^{2}\text{ ,}
\end{equation}
where $\hat{y}^{l}, \hat{y}_{0}^{l} \in \mathbb{R}^{H_{l} \times W_{l} \times C_{l}}$ are the channel-dimension unit-normalized $l$-th layer feature stack of rendered and reference image patches, which is extracted from $L$ layers of a fixed pre-trained neural network such as VGG~\cite{vgg}. Note that to use the perceptual loss we need to sample rays in an image patch level.

\begin{figure}[!tb]
  \centering
  \includegraphics[width=\linewidth]{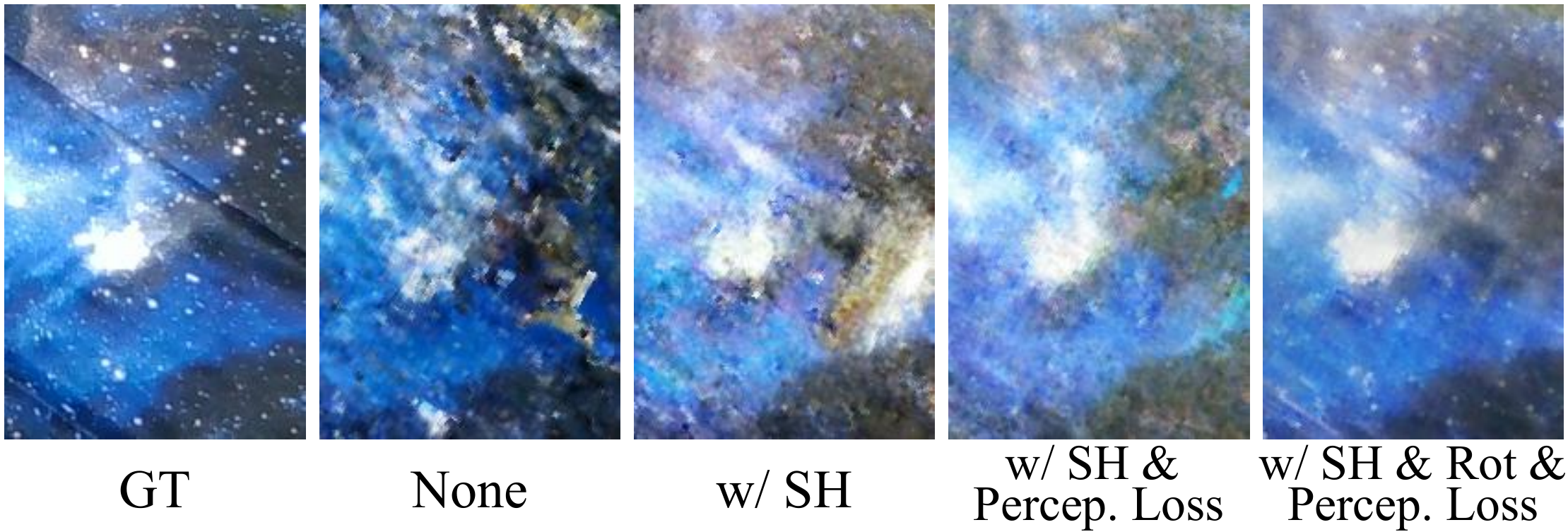}
  \caption{\textbf{Effectiveness of SH, perception loss and view augmentation through random rotation on novel view.} We can see that all of them can markedly enhance the synthesis quality.}
  \vspace{-0.1in}
  \label{fig:percep_rot}
\end{figure}

Combining the above, we can get an overall loss:
\begin{equation}
\mathcal{L}_{\mathrm{overall}}=\mathcal{L}_{\mathrm{render}}+\lambda_{\mathrm{percep}}\mathcal{L}_{\mathrm{percep}}\text{ ,}
\end{equation}
where $\lambda_{\mathrm{percep}}$ is a weighting hyper-parameter.

Lastly, as mentioned in Sec.~\ref{sec:model_architecture}, we have multi-stage outputs, so the final total loss is: % the weighted sum of the loss in different stages:
\begin{equation}
\mathcal{L}_{\mathrm{total}}=\sum_s \lambda_{\mathrm{stage}}^s \mathcal{L}_{\mathrm{overall}}^s\text{ ,}
\end{equation}
where $\lambda_{\mathrm{stage}}^s$ is the weight coefficient of stage $s$.

\subsection{View Augmentation}
\label{sec:view_aug}
We utilize spherical harmonics to make rendered colors view-independent (see Sec.~\ref{sec:explicit_modeling}). Nevertheless, due to the sparsity and small quantity of training views, we observe in the experiments that the SH coefficients are not fitted well on novel view directions. To handle this limitation, we apply random rotation augmentation on input point clouds to manually simulate unseen view directions. We find that this simple operation can significantly improve the quality of rendered unseen view images. Fig.~\ref{fig:percep_rot} shows the effectiveness of implementing SH, perception loss and view augmentation through random rotation on point clouds.

\subsection{Fast Inference}
% Many previous works such as~\cite{dvgo, nsvf} utilize explicit structures to gain super fast inference process. Our fully explicit approach also boast the fast inference ability.
Most fully implicit coordinate-based NeRF methods struggle with the rendering efficiency because they have to run networks repeatedly on each position on each ray of given camera poses. However, X-NeRF can just save the explicit scene representations $\mathscr{T}_{\mathrm{out}}$ so that only rendering operations such as interpolation and integral that are highly parallelizable needed to be done during inference. The inference complexity is thus reduced a lot since no neural network is needed to be run. % It typically can bring tens or even hundreds of times acceleration.

\section{Experiments}
\label{sec:exp}

\begin{figure*}[!tb]
  \centering
  \includegraphics[width=0.95\linewidth]{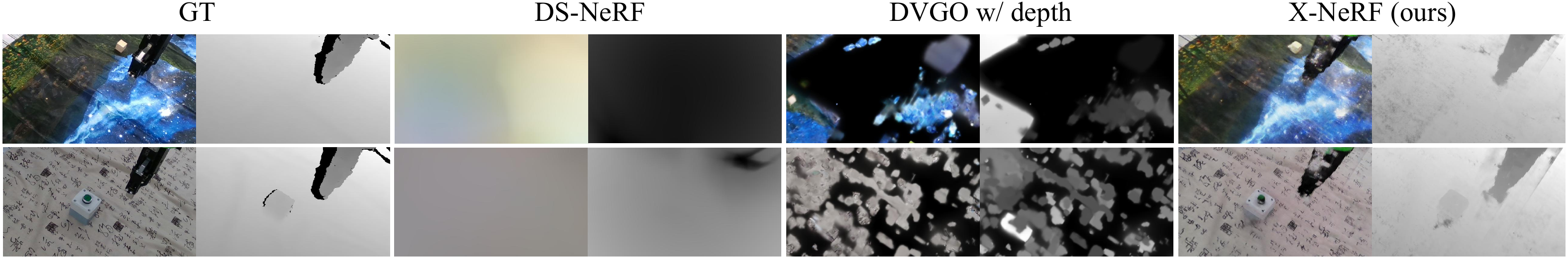}
  \caption{\textbf{Single scene qualitative results on scene 1-2.} From top to bottom, each line shows rendered RGB and depth images on novel view of different methods. Obviously, our proposed X-NeRF performs significantly better than the other two implicit methods.}
%   \vspace{-0.1in}
  \label{fig:single_comparison}
\end{figure*}

\begin{table*}[th]
\begin{center}
\resizebox{0.95\linewidth}{!}
{
\begin{tabular}{l|cccc|cccc|cccc}
\toprule
\multirow{3}{*}{} & \multicolumn{4}{c|}{Scene 1}                                                                                                                                   & \multicolumn{4}{c|}{Scene 2}                                                                                                                                   & \multicolumn{4}{c}{Scene 3}                                                                                                                                    \\ \cline{2-13} 
                  & \multicolumn{3}{c|}{RGB Metrics}                                         & \multirow{2}{*}{\begin{tabular}[c]{@{}c@{}}Depth \\ Err\%$\downarrow$\end{tabular}} & \multicolumn{3}{c|}{RGB Metrics}                                         & \multirow{2}{*}{\begin{tabular}[c]{@{}c@{}}Depth \\ Err\%$\downarrow$\end{tabular}} & \multicolumn{3}{c|}{RGB Metrics}                                          & \multirow{2}{*}{\begin{tabular}[c]{@{}c@{}}Depth \\ Err\%$\downarrow$\end{tabular}} \\ \cline{2-4} \cline{6-8} \cline{10-12}
                  & LPIPS$\downarrow$ & PSNR$\uparrow$ & \multicolumn{1}{c|}{SSIM$\uparrow$} &                                                                                     & LPIPS$\downarrow$ & PSNR$\uparrow$ & \multicolumn{1}{c|}{SSIM$\uparrow$} &                                                                                     & LPIPS$\downarrow$ & PSNR$\uparrow$ & \multicolumn{1}{c|}{SSIM$\uparrow$}                      &                                                                                     \\ \midrule
DS-NeRF~\cite{dsnerf}           & 0.891             & 6.65           & 0.267                               & 87.54                                                                               & 0.714             & 15.60          & \textbf{0.537}                      & 86.55                                                                               & 0.797             & 8.02           & 0.011                               & 82.85                                                                               \\
DVGO~\cite{dvgo}              & 0.735             & 8.93           & 0.100                               & 68.46                                                                               & 0.738             & 9.47           & 0.205                               & 55.13                                                                               & 0.776             & 9.62           & 0.156                               & 69.81                                                                               \\
DVGO~\cite{dvgo} w/ depth     & 0.726             & 9.46          & 0.124                               & 66.16                                                                               & 0.723             & 10.03          & 0.220                               & 55.69                                                                               & 0.764             & 10.22          & 0.170                               & 69.21                                                                               \\
X-NeRF            & \textbf{0.521}    & \textbf{17.39} & \textbf{0.414}                      & \textbf{0.257}                                                                      & \textbf{0.505}    & \textbf{16.38} & 0.457                               & \textbf{0.356}                                                                      & \textbf{0.452}    & \textbf{17.83} & \textbf{0.477}                      & \textbf{1.66}                                                                       \\ \midrule\midrule
\multirow{3}{*}{} & \multicolumn{4}{c|}{Scene 4}                                                                                                                                   & \multicolumn{4}{c|}{Scene 5}                                                                                                                                   & \multicolumn{4}{c}{Scene 6}                                                                                                                                    \\ \cline{2-13} 
                  & \multicolumn{3}{c|}{RGB Metrics}                                         & \multirow{2}{*}{\begin{tabular}[c]{@{}c@{}}Depth \\ Err\%$\downarrow$\end{tabular}} & \multicolumn{3}{c|}{RGB Metrics}                                         & \multirow{2}{*}{\begin{tabular}[c]{@{}c@{}}Depth \\ Err\%$\downarrow$\end{tabular}} & \multicolumn{3}{c|}{RGB Metrics}                                         & \multirow{2}{*}{\begin{tabular}[c]{@{}c@{}}Depth \\ Err\%$\downarrow$\end{tabular}} \\ \cline{2-4} \cline{6-8} \cline{10-12}
                  & LPIPS$\downarrow$ & PSNR$\uparrow$ & \multicolumn{1}{c|}{SSIM$\uparrow$} &                                                                                     & LPIPS$\downarrow$ & PSNR$\uparrow$ & \multicolumn{1}{c|}{SSIM$\uparrow$} &                                                                                     & LPIPS$\downarrow$ & PSNR$\uparrow$ & \multicolumn{1}{c|}{SSIM$\uparrow$} &                                                                                     \\ \midrule
DS-NeRF~\cite{dsnerf}           & 0.698             & 7.07           & 0.093                               & 87.56                                                                               & 0.701             & 12.00          & 0.748                               & 86.14                                                                               & 0.754             & 8.63           & 0.463                               & 80.04                                                                               \\
DVGO~\cite{dvgo}              & 0.651             & 11.84          & 0.576                               & 51.67                                                                               & 0.729             & 11.89          & 0.540                               & 58.10                                                                               & 0.740             & 7.36           & 0.226                               & 59.71                                                                               \\
DVGO~\cite{dvgo} w/ depth     & 0.643             & 11.96          & 0.560                               & 54.69                                                                               & 0.730             & 12.05          & 0.532                               & 58.38                                                                               & 0.762             & 7.51           & 0.196                               & 61.51                                                                               \\
X-NeRF            & \textbf{0.397}    & \textbf{17.73} & \textbf{0.754}                      & \textbf{0.367}                                                                      & \textbf{0.431}    & \textbf{18.58} & \textbf{0.812}                      & \textbf{0.485}                                                                      & \textbf{0.471}    & \textbf{18.19} & \textbf{0.593}                      & \textbf{0.269}                                                                      \\ \bottomrule
\end{tabular}
}
\end{center}
\caption{\textbf{Quantitative results on each single scene.} We use three common RGB metrics, namely LPIPS (using pre-trained VGG, lower is better) and PSNR/SSIM(higher is better). The depth error is evaluated by mean squared error in valid area, whose unit is meter\%.}
\label{tab:single_comparison}
\end{table*}

\subsection{Dataset}

Since the setting of multi-scene insufficient 360$^{\circ} $ RGB-D views has never been discussed before, we collect a new dataset for this challenging task. We use 7 RGB-D cameras to capture 6 seen and 4 novel scenes, in which a robot arm is doing different tasks in different environments. The views are extremely sparse with large angle transformations. The overlapping among views is less than 10\% to 20\%. One example is shown in Fig.~\ref{fig:insufficient_views}. Among 7 views, one is left for testing while the other 6 are training views. Please refer to the supplementary material for more details.

\subsection{Implementation Details}
All experiments are conducted on a single NVIDIA A100 GPU. We apply PyTorch~\cite{pytorch} and Minkowski Engine~\cite{me} to build our sparse network. Among all single scene experiments and the multi-scene experiment, we keep the same hyper-parameters. A simple ResNet14~\cite{resnet} of 3D sparse version is employed as our backbone. When voxelizing the input point clouds, we set the voxel size as $4\times 10^{-3}$. We choose AdamW~\cite{adamw} as our optimizer. In each batch, we sample 2 random image patch of size $40\times 40$ for all 6 training views, which is equivalent to a total ray batch size of $6\times 2 \times 40 \times 40=19200$. We train our models for $240$ epochs with an initial learning rate of $10^{-3}$, and the learning rate is divided by $10$ at $120$th and $200$th epoch. See supplementary material for detailed hyper-parameter setups.

\subsection{Comparison Experiments}
In this section, we compare proposed X-NeRF with state-of-the-art implicit NeRF-related work on our extremely challenging 360$^{\circ} $ insufficient RGB-D view dataset. Note that besides common metrics for RGB novel view synthesis, we also evaluate depth error since in our setting of insufficient RGB-D input views, the rendering quality of depth counts for much. We adopt mean squared error in meters in valid areas as depth metric since low-cost depth cameras may have some invalid values.

{\bf Single Scene Comparisons.}
Considering our data and models are RGB-D, we first compare with DS-NeRF~\cite{dsnerf}, a state-of-the-art implicit NeRF-based method that also allows depth inputs. Furthermore, we also compare with DVGO~\cite{dvgo}, a state-of-the-art NeRF-based approach that utilizes explicit voxel grid structures. Unfortunately, the above two methods do not support multi-scene training, so we compare with them on single scene. DVGO~\cite{dvgo} originally does not support depth supervision. For fair comparison, we also add depth supervision to it. The quantitative metrics on novel view can be found in Tab.~\ref{tab:single_comparison}.  We can see that X-NeRF significantly out-performs the two methods on each single scene especially on depth error, which means that X-NeRF succeeds in avoiding to overfit to a trivial solution. Moreover, DVGO~\cite{dvgo} does not have a significant improvement on novel view after adding depth supervision, indicating that the key factor is not the depth loss but the modeling methodology. The qualitative results of scene 1-3 are displayed in Fig.~\ref{fig:single_comparison} and all results can be found in supplementary material. It is obvious that implicit methods fails to generalize well on novel view facing insufficient training views and large view gaps. % , while proposed explicit approach can deal with it robustly.

{\bf Multi-Scene and Cross-Scene Comparisons.}
As mentioned above, X-NeRF is able to deal with multi-scene representation due to our explicit completion modeling. There is a little work that can handle multi-scene task. PixelNeRF~\cite{pixel_nerf} combines image features from 2D CNNs with NeRF so that it can train on multi-scene. Therefore, we re-train pixelNeRF~\cite{pixel_nerf} concurrently on 6 scenes for $3000$ epochs to compare our work with it on the multi-scene performance. IBRNet~\cite{ibrnet} is an image-based
rendering approach which applies an MLP and a ray transformer to learn a generic
view interpolation function. We finetune IBRNet~\cite{ibrnet} using its pre-trained weights for 60000 iterations. Since the two methods both need reference views when rendering, we use all the 6 seen views as reference images during evaluation. From the quantitative results in Tab.~\ref{tab:multi_comparison} and the qualitative results of scene 3-6 in Fig.~\ref{fig:multi_comparison} (all results can be found in supplementary material), we can see that pixelNeRF~\cite{pixel_nerf} completely overfit to a trivial solution, and X-NeRF again beat both the two methods. If we compare Tab.~\ref{tab:multi_comparison} with Tab.~\ref{tab:single_comparison}, the multi-scene version X-NeRF is even better than the single-scene version, indicating that X-NeRF has a powerful generalization capacity. We also do cross-scene comparisons on novel view of two novel scenes. The results indicate that X-NeRF has a powerful cross-scene performance, which proves that X-NeRF learns a completion mapping with good generalization ability.

\begin{figure}[tb]
\begin{center}
   \includegraphics[width=\linewidth]{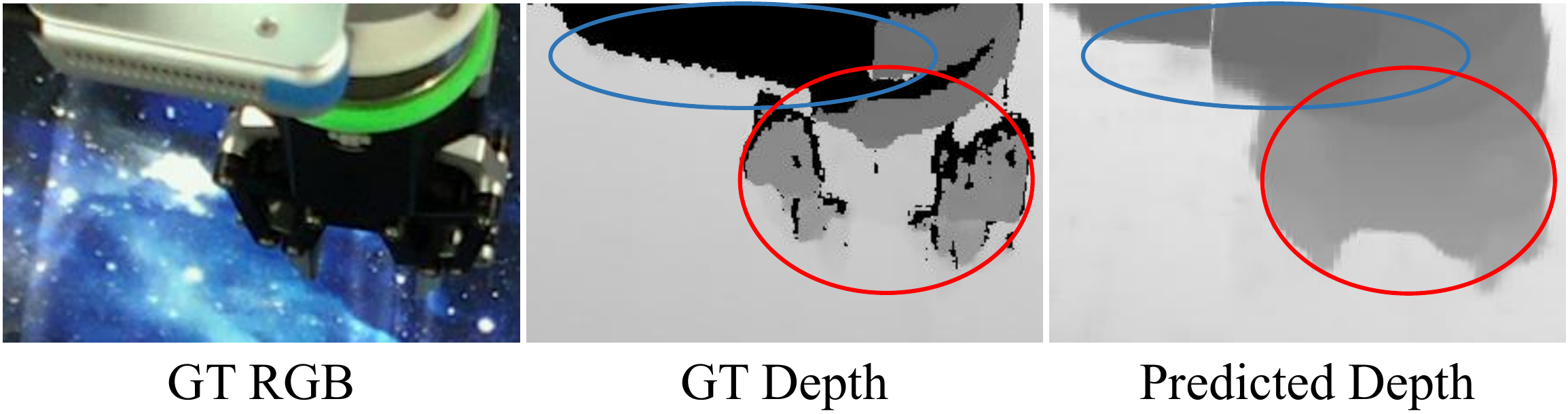}
\end{center}
\vspace{-0.1in}
   \caption{\textbf{Depth completion effect.} Examples of ground truth RGB image, ground truth depth image and predicted depth image are shown from left to right. The invalid depth values are represented in black color and marked with a blue ellipse. The red ellipse circles out the area where the depth camera has errors. We can see that the rendering result can not only complete the missing area but can also correct the mistakes.}
\vspace{-0.1in}
\label{fig:completion}
\end{figure}

\begin{figure*}[htb]
  \centering
  \includegraphics[width=0.98\linewidth]{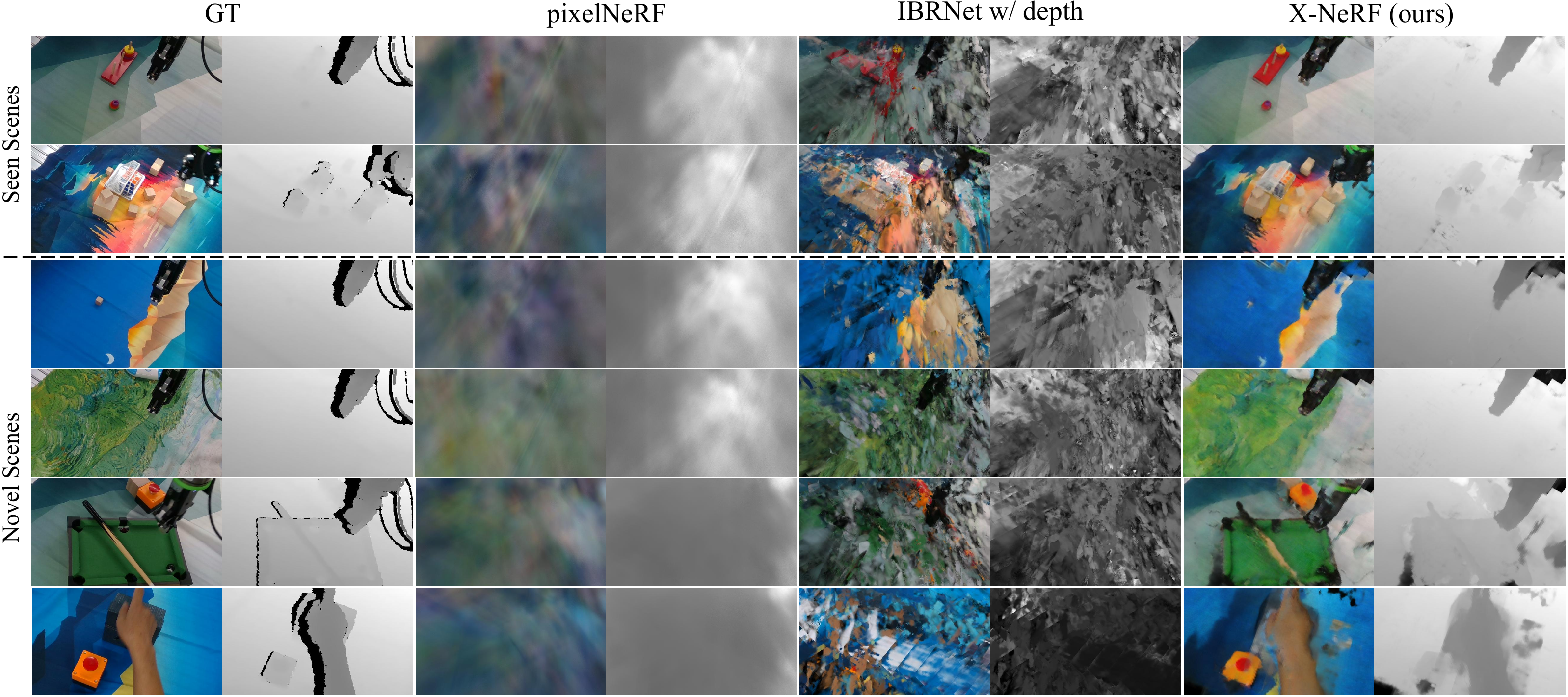}
  \caption{\textbf{Multi-scene and cross-scene qualitative results.} The first 2 rows show scene 5-6 and the last 4 rows show the 4 novel scenes.}
  \vspace{-0.05in}
  \label{fig:multi_comparison}
\end{figure*}

\begin{table*}[htb]
\begin{center}
\resizebox{\linewidth}{!}
{
\begin{tabular}{l|cccc|cccc|cccc|cccc}
\toprule
\multirow{3}{*}{} & \multicolumn{4}{c|}{Scene 1}                                                                                                                                   & \multicolumn{4}{c|}{Scene 2}                                                                                                                                   & \multicolumn{4}{c|}{Scene 3}                                                                                                                                   & \multicolumn{4}{c}{Scene 4}                                                                                                                                    \\ \cline{2-17} 
                  & \multicolumn{3}{c|}{RGB Metrics}                                         & \multirow{2}{*}{\begin{tabular}[c]{@{}c@{}}Depth \\ Err\%$\downarrow$\end{tabular}} & \multicolumn{3}{c|}{RGB Metrics}                                         & \multirow{2}{*}{\begin{tabular}[c]{@{}c@{}}Depth \\ Err\%$\downarrow$\end{tabular}} & \multicolumn{3}{c|}{RGB Metrics}                                         & \multirow{2}{*}{\begin{tabular}[c]{@{}c@{}}Depth \\ Err\%$\downarrow$\end{tabular}} & \multicolumn{3}{c|}{RGB Metrics}                                         & \multirow{2}{*}{\begin{tabular}[c]{@{}c@{}}Depth \\ Err\%$\downarrow$\end{tabular}} \\ \cline{2-4} \cline{6-8} \cline{10-12} \cline{14-16}
                  & LPIPS$\downarrow$ & PSNR$\uparrow$ & \multicolumn{1}{c|}{SSIM$\uparrow$} &                                                                                     & LPIPS$\downarrow$ & PSNR$\uparrow$ & \multicolumn{1}{c|}{SSIM$\uparrow$} &                                                                                     & LPIPS$\downarrow$ & PSNR$\uparrow$ & \multicolumn{1}{c|}{SSIM$\uparrow$} &                                                                                     & LPIPS$\downarrow$ & PSNR$\uparrow$ & \multicolumn{1}{c|}{SSIM$\uparrow$} &                                                                                     \\ \midrule
pixelNeRF~\cite{pixel_nerf}         & 0.802             & 11.72          & 0.333                               & 25.76                                                                               & 0.792             & 16.75          & 0.507                               & 22.45                                                                               & 0.785             & 13.14          & \multicolumn{1}{c|}{0.359}          & 25.59                                                                               & 0.770             & 12.68          & \multicolumn{1}{c|}{0.576}          & 21.71                                                                               \\
IBRNet~\cite{ibrnet}            & 0.646             & 14.37         & 0.283                               & 9.30                                                                                & 0.673             & 16.43          & 0.316                               & 12.42                                                                               & 0.631             & 16.89          & \multicolumn{1}{c|}{0.336}          & 12.38                                                                               & 0.642             & 13.95          & \multicolumn{1}{c|}{0.533}          & 8.31                                                                                \\
IBRNet~\cite{ibrnet} w/ depth   & 0.620             & 14.85          & 0.336                               & 11.48                                                                               & 0.657             & 16.63          & 0.368                               & 16.79                                                                               & 0.617             & 16.65          & \multicolumn{1}{c|}{0.351}          & 12.05                                                                               & 0.631             & 14.67          & \multicolumn{1}{c|}{0.552}          & 8.96                                                                                \\
X-NeRF (ours)            & \textbf{0.534}    & \textbf{18.25} & \textbf{0.440}                      & \textbf{0.0971}                                                                     & \textbf{0.587}    & \textbf{17.59} & \textbf{0.520}                      & \textbf{0.144}                                                                      & \textbf{0.484}    & \textbf{18.05} & \multicolumn{1}{c|}{\textbf{0.476}} & \textbf{0.888}                                                                      & \textbf{0.397}    & \textbf{18.65} & \multicolumn{1}{c|}{\textbf{0.752}} & \textbf{0.138}                                                                      \\ \midrule
\multirow{3}{*}{} & \multicolumn{4}{c|}{Scene 5}                                                                                                                                   & \multicolumn{4}{c|}{Scene 6}                                                                                                                                   & \multicolumn{4}{c|}{Novel Scene 1}                                                                                                                             & \multicolumn{4}{c}{Novel Scene 2}                                                                                                                              \\ \cline{2-17} 
                  & \multicolumn{3}{c|}{RGB Metrics}                                         & \multirow{2}{*}{\begin{tabular}[c]{@{}c@{}}Depth \\ Err\%$\downarrow$\end{tabular}} & \multicolumn{3}{c|}{RGB Metrics}                                         & \multirow{2}{*}{\begin{tabular}[c]{@{}c@{}}Depth \\ Err\%$\downarrow$\end{tabular}} & \multicolumn{3}{c|}{RGB Metrics}                                         & \multirow{2}{*}{\begin{tabular}[c]{@{}c@{}}Depth \\ Err\%$\downarrow$\end{tabular}} & \multicolumn{3}{c|}{RGB Metrics}                                         & \multirow{2}{*}{\begin{tabular}[c]{@{}c@{}}Depth \\ Err\%$\downarrow$\end{tabular}} \\ \cline{2-4} \cline{6-8} \cline{10-12} \cline{14-16}
                  & LPIPS$\downarrow$ & PSNR$\uparrow$ & \multicolumn{1}{c|}{SSIM$\uparrow$} &                                                                                     & LPIPS$\downarrow$ & PSNR$\uparrow$ & \multicolumn{1}{c|}{SSIM$\uparrow$} &                                                                                     & LPIPS$\downarrow$ & PSNR$\uparrow$ & \multicolumn{1}{c|}{SSIM$\uparrow$} &                                                                                     & LPIPS$\downarrow$ & PSNR$\uparrow$ & \multicolumn{1}{c|}{SSIM$\uparrow$} &                                                                                     \\ \midrule
pixelNeRF~\cite{pixel_nerf}         & 0.707             & 14.03          & 0.714                               & 24.28                                                                               & 0.808             & 11.72          & 0.477                               & 19.62                                                                               & 0.885             & 11.74          & \multicolumn{1}{c|}{0.546}          & 23.02                                                                               & 0.722             & 15.22          & \multicolumn{1}{c|}{0.388}          & 24.24                                                                               \\
IBRNet~\cite{ibrnet}            & 0.633             & 18.49          & 0.613                               & 8.83                                                                                & 0.656             & 13.45          & 0.365                               & 5.38                                                                                & 0.633             & 14.06          & \multicolumn{1}{c|}{0.455}          & 9.42                                                                                & 0.638             & 16.33          & \multicolumn{1}{c|}{0.294}          & 12.36                                                                               \\
IBRNet~\cite{ibrnet} w/ depth   & 0.678             & 14.52          & 0.512                               & 12.68                                                                               & 0.649             & 13.74          & 0.356                               & 5.96                                                                                & 0.628             & 13.55          & \multicolumn{1}{c|}{0.463}          & 8.72                                                                                & 0.648             & 14.36          & \multicolumn{1}{c|}{0.261}          & 14.67                                                                               \\
X-NeRF (ours)            & \textbf{0.408}    & \textbf{18.71} & \textbf{0.817}                      & \textbf{0.346}                                                                      & \textbf{0.476}    & \textbf{17.69} & \textbf{0.605}                      & \textbf{0.152}                                                                      & \textbf{0.452}    & \textbf{18.66} & \multicolumn{1}{c|}{\textbf{0.603}} & \textbf{1.70}                                                                       & \textbf{0.549}    & \textbf{17.93} & \multicolumn{1}{c|}{\textbf{0.445}} & \textbf{1.82}                                                                       \\ \midrule
\multirow{3}{*}{} & \multicolumn{4}{c|}{Novel Scene 3}                                                                                                                             & \multicolumn{4}{c|}{Novel Scene 4}                                                                                                                             & \multicolumn{4}{c|}{Overall Avg. on Seen Scenes}                                                                                                               & \multicolumn{4}{c}{Overall Avg. on Novel Scenes}                                                                                                               \\ \cline{2-17} 
                  & \multicolumn{3}{c|}{RGB Metrics}                                         & \multirow{2}{*}{\begin{tabular}[c]{@{}c@{}}Depth \\ Err\%$\downarrow$\end{tabular}} & \multicolumn{3}{c|}{RGB Metrics}                                         & \multirow{2}{*}{\begin{tabular}[c]{@{}c@{}}Depth \\ Err\%$\downarrow$\end{tabular}} & \multicolumn{3}{c|}{RGB Metrics}                                         & \multirow{2}{*}{\begin{tabular}[c]{@{}c@{}}Depth \\ Err\%$\downarrow$\end{tabular}} & \multicolumn{3}{c|}{RGB Metrics}                                         & \multirow{2}{*}{\begin{tabular}[c]{@{}c@{}}Depth \\ Err\%$\downarrow$\end{tabular}} \\ \cline{2-4} \cline{6-8} \cline{10-12} \cline{14-16}
                  & LPIPS$\downarrow$ & PSNR$\uparrow$ & \multicolumn{1}{c|}{SSIM$\uparrow$} &                                                                                     & LPIPS$\downarrow$ & PSNR$\uparrow$ & \multicolumn{1}{c|}{SSIM$\uparrow$} &                                                                                     & LPIPS$\downarrow$ & PSNR$\uparrow$ & \multicolumn{1}{c|}{SSIM$\uparrow$} &                                                                                     & LPIPS$\downarrow$ & PSNR$\uparrow$ & \multicolumn{1}{c|}{SSIM$\uparrow$} &                                                                                     \\ \midrule
pixelNeRF~\cite{pixel_nerf}         & 0.679             & 14.57          & 0.630                               & 28.72                                                                               & 0.748             & 13.24          & 0.580                               & 22.74                                                                               & 0.778             & 13.34          & \multicolumn{1}{c|}{0.494}          & 23.23                                                                               & 0.759             & 13.69          & \multicolumn{1}{c|}{0.536}          & 24.68                                                                               \\
IBRNet~\cite{ibrnet}            & 0.692             & 13.44          & 0.418                               & 7.26                                                                                & 0.713             & 11.95          & 0.340                               & 10.45                                                                               & 0.647             & 15.60          & \multicolumn{1}{c|}{0.408}          & 9.44                                                                                & 0.669             & 13.95          & \multicolumn{1}{c|}{0.377}          & 9.87                                                                                \\
IBRNet~\cite{ibrnet} w/ depth   & 0.696             & 13.60          & 0.379                               & 10.17                                                                               & 0.724             & 11.50          & 0.320                               & 13.62                                                                               & 0.642             & 15.18          & \multicolumn{1}{c|}{0.412}          & 11.32                                                                               & 0.674             & 13.25          & \multicolumn{1}{c|}{0.356}          & 11.80                                                                               \\
X-NeRF (ours)            & \textbf{0.512}    & \textbf{17.17} & \textbf{0.656}                      & \textbf{1.70}                                                                       & \textbf{0.520}    & \textbf{17.57} & \textbf{0.633}                      & \textbf{1.67}                                                                       & \textbf{0.486}    & \textbf{18.19} & \multicolumn{1}{c|}{\textbf{0.582}} & \textbf{0.661}                                                                      & \textbf{0.508}    & \textbf{17.83} & \multicolumn{1}{c|}{\textbf{0.584}} & \textbf{1.72}                                                                       \\ \bottomrule
\end{tabular}
}
\end{center}
\caption{\textbf{Quantitative results on multi-scene.} We train each model on 6 scenes simultaneously, then report their performance on each scene as well as the overall average scores. We also evaluate the models on two novel scenes on novel view synthesis.}
\label{tab:multi_comparison}
\end{table*}

{\bf Complexity Comparisons.}
We further compare the inference time per image, training time and the model size among the multi-scene methods, which is shown in Tab.~\ref{tab:complexity}. We can find that X-NeRF requires less training time. X-NeRF also has a comparable inference time and model size with IBRNet~\cite{ibrnet}, much better than pixelNeRF~\cite{pixel_nerf}.

\begin{table}[bt]
\begin{center}
\resizebox{0.85\linewidth}{!}{
\begin{tabular}{l|ccc}
\toprule
Method    & \#Params & Training       & Inference \\ \midrule
pixelNeRF~\cite{pixel_nerf} & 28.2M    & $\textgreater$10 days  & $\sim$41s \\
IBRNet~\cite{ibrnet}    & \textbf{8.9M}     & $\sim$6 days   & $\sim$18s \\
X-NeRF (ours)      & 22.1M    & \textbf{$\sim$3.5 days} & \textbf{$\sim$11s} \\ \bottomrule
\end{tabular}}
\end{center}
\caption{\textbf{Complexity comparisons.} We report model size, training time (including pre-training) and inference time per image.}
\label{tab:complexity}
\end{table}

\subsection{Completion Ability}
We further study X-NeRF's completion ability. 
% As mentioned above, low-cost depth cameras have inevitable invalid and wrong areas, as shown in Fig.~\ref{fig:completion}. 
The result in Fig.~\ref{fig:completion} clearly illustrates that X-NeRF are robust enough to complete the missing areas and correct the wrong values caused by low-cost depth cameras.

% \subsection{Comparisons on Multi-Scene}

\section{Conclusion}
We propose a novel and extremely challenging task that to synthesize novel view RGB-D images given only insufficient seen views on multiple scenes. The problem is useful in low-cost environment and may expand practical usage scenarios of NeRF-related work. A new dataset is collected for the new problem, on which our proposed fully explicit methodology X-NeRF greatly out-performs existing methods. % X-NeRF may have more potential if trained on massive scenes which we leave for future work.

\section*{Acknowledgement}
Thanks for the selfless help and valuable advice from Hao-Shu Fang and Prof. Cewu Lu!

{\small
\bibliographystyle{ieee_fullname}
\bibliography{egbib}
}

\end{document}